\documentclass[sigconf,screen]{acmart}
\AtBeginDocument{%
  }

\setcopyright{none}
\copyrightyear{2026}
\acmYear{2026}
\acmDOI{XXXXXXX.XXXXXXX}
\acmBooktitle{Companion Proceedings of the 34th ACM Symposium on the Foundations of Software Engineering (FSE '26), June 5--9, 2026, Montreal, Canada}
\acmISBN{978-1-4503-XXXX-X/2018/06}

\usepackage{listings}
\usepackage{xcolor}
\usepackage[utf8]{inputenc}
\usepackage[T1]{fontenc}

\lstdefinestyle{acmcode}{
  basicstyle=\ttfamily\footnotesize,
  keywordstyle=\color{blue},
  commentstyle=\color{gray},
  stringstyle=\color{teal},
  numbers=left,
  numberstyle=\tiny,
  stepnumber=1,
  numbersep=5pt,
  frame=single,
  breaklines=true,
  breakatwhitespace=true,
  showstringspaces=false,
  tabsize=2,
  captionpos=b
}

\begin{document}

\title{Panther: Faster and Cheaper Computations with Randomized Numerical Linear Algebra}

\author{Fahd Seddik}
\authornote{Authors contributed equally to this research.}
\authornote{Work done while the authors were undergraduate students at Cairo University.}
\affiliation{
  \institution{University of British Columbia}
  \city{Kelowna}
  \state{BC}
  \country{Canada}
}
\email{fahd.seddik@ubc.ca}

\author{Abdulrahman Elbedewy}
\authornotemark[1]
\authornotemark[2]
\affiliation{
  \institution{University of Texas at Austin}
  \city{Austin}
  \state{Texas}
  \country{USA}
}
\email{abdulrahman.elbedewy@utexas.edu}

\author{Gaser Sami}
\authornotemark[1]
\authornotemark[2]
\affiliation{
  \institution{Cairo University}
  \city{Giza}
  \country{Egypt}
}
\email{gaser.elmasry02@eng-st.cu.edu.eg}

\author{Mohamed Abdelmoniem}
\authornotemark[1]
\authornotemark[2]
\affiliation{
  \institution{Noon}
  \city{Cairo}
  \country{Egypt}
}
\email{moabdelmoniem@noon.com}

\author{Yahia Zakaria}
\affiliation{
  \institution{Cairo University}
  \city{Giza}
  \country{Egypt}
}
\email{yzetman@eng.cu.edu.eg}

\renewcommand{\shortauthors}{Seddik et al.}

\begin{abstract}
Training modern deep learning models is increasingly constrained by GPU memory and compute limits. While Randomized Numerical Linear Algebra (RandNLA) offers proven techniques to compress these models, the lack of a unified, production-grade library prevents widely adopting these methods. We present Panther, a PyTorch-compatible library that consolidates established RandNLA algorithms into a single high-performance framework. Panther engineers efficient, drop-in replacements for standard components including sketched linear layers, 2D convolution, multi-head attention, and randomized matrix decompositions (such as pivoted CholeskyQR). By implementing a custom C++/CUDA backend (pawX), Panther provides an optimized implementation that can run on both CPUs and GPUs. We demonstrate the effectiveness of RandNLA techniques and Panther's ease of adoption. By replacing standard PyTorch linear layers with Panther layers (requiring only a few lines of code) we achieve significant memory savings (up to 75\%) on BERT while maintaining comparable loss. Source code is available (MIT License) at \url{https://github.com/FahdSeddik/panther}, along with demonstration video at \url{https://youtu.be/7M3RQb4KWxs}.
\end{abstract}




\maketitle

\section{Introduction}
\label{sec:intro}

The rapid growth of neural network models into the billions of parameters has made memory footprint and computational cost central bottlenecks for both research and deployment \cite{zhen-etal-2025-taming, zhu2024survey}. Core building blocks such as linear layers, convolutions, and attention mechanisms rely heavily on dense matrix operations whose time and space complexity scale poorly with model size. As a result, training and inference increasingly demand specialized hardware and large GPU budgets, limiting accessibility for researchers and complicating deployment on resource-constrained platforms.

Randomized numerical linear algebra (RandNLA) provides a principled family of techniques—random projection, sketching, and randomized matrix factorizations—that reduce arithmetic and memory costs while offering probabilistic approximation guarantees. Over the past decade, algorithms such as randomized singular value decomposition (RSVD), sketching-based regression, and randomized QR variants have matured into well-understood tools with strong theoretical foundations and growing empirical validation \cite{melnichenko2025choleskyqrrandomizationpivotingtall, murray2023randomizednumericallinearalgebra}. Despite this progress, most RandNLA methods remain difficult to use in practice: existing implementations are often present in disparate repositories and across different frameworks, creating a substantial gap between theory and deployable systems. Although introduced in \cite{murray2023randomizednumericallinearalgebra}, RandBLAS and RandLAPACK have not been adopted by mainstream machine learning libraries (for example, PyTorch).

We present \textbf{Panther}, a PyTorch-oriented library that bridges this gap by bringing production-quality RandNLA theory into standard machine learning workflows. Panther provides drop-in replacements for common PyTorch layers, including linear layers, 2D convolutions following the work of \cite{kasiviswanathan2017deepneuralnetworkapproximation}, and multi-head attention based on random-feature approximations \cite{choromanski2022rethinkingattentionperformers} all while maintaing full integration with PyTorch and similar APIs to avoid major refactoring work. At the algorithmic level, Panther implements core randomized decompositions such as RSVD and CholeskyQR with randomized pivoting for tall matrices (CQRRPT) \cite{melnichenko2025choleskyqrrandomizationpivotingtall}, following best practices established in the RandNLA literature for numerical stability and accuracy.

Panther is designed with both usability and performance in mind. Its three-layer architecture comprising a Python-facing API, Python bindings, and a native C++/CUDA backend allowing users to replace exact layers with randomized counterparts using only a few lines of code, while retaining autograd support and GPU acceleration with PyTorch. To reduce the burden of selecting extra sketching hyperparameters that are introduced by RandNLA, Panther includes an Optuna-based \cite{ozaki2025optunahub} AutoTuner that automatically searches for configurations meeting user-specified accuracy and resource constraints.

By packaging theoretically grounded RandNLA algorithms into a practical, developer-friendly tool, Panther enables systematic exploration of approximation--efficiency trade-offs in large neural networks. This paper demonstrates how Panther lowers the barrier to adopting RandNLA directly into PyTorch models and supports both research experimentation and production deployment.
\section{Panther Design}
\label{sec:method}

\subsection{Architecture and Core Engine}
\label{subsec:arch}
At the user level, Panther provides a Python API. However, these components delegate heavy lifting to the bottom tier, \texttt{pawX}, a performance core written as a PyTorch extension. This enables all operations to be directly integrated with ATen \cite{Ansel_PyTorch_2_Faster_2024}. Our backend, \texttt{pawX}, bundles OpenBLAS and employs custom CUDA kernels that utilize NVIDIA Tensor Cores via the Warp Matrix Multiply-Accumulate (WMMA) API. We adopt the mathematical formulation for Linear and Conv2D sketching from \cite{kasiviswanathan2017deepneuralnetworkapproximation}, while our randomized linear attention mechanism aligns with the framework proposed for Performers \cite{choromanski2022rethinkingattentionperformers}.

\subsection{AutoTuner Module}
\label{subsec:tuning}
Selecting optimal sketching parameters is a significant barrier to adopting RandNLA. Panther addresses this via the \texttt{tuner} module, which includes the \texttt{SKAutoTuner} built on Optuna~\cite{ozaki2025optunahub}. Users specify high-level constraints, such as a memory budget or accuracy tolerance, and the tuner explores the configuration space. This automates the trade-off analysis between speed, memory, and accuracy, allowing practitioners to utilize Panther without requiring deep expertise in RandNLA.

The tuner module also serves as a way for users who want to adopt Panther to easily integrate it into their existing workflows. SKAutoTuner can be given a torch-saved model provided with regex or specific layers to replace and it automatically figures out the optimum extra hyperparameters that sketching introduces.

\section{Tool Usage}
\label{sec:tool}

Panther prioritizes ease of access, requiring only a standard pip installation. Users with CUDA 12.4-enabled GPUs on Windows can install via PyPi, while CPU-only systems or Linux users build from source using the provided instructions in the repository.\footnote{A docker image is provided via \texttt{docker pull fahdseddik/panther-dev}.}

\subsection{During Development use-case}

The API is designed to be a drop-in replacement for torch.nn, minimizing code refactoring. Converting a standard PyTorch model requires only a single line change per layer: \lstinline|Linear(8192, 8192)| becomes  \lstinline|SKLinear(8192, 8192, num_terms=1, low_rank=16)|. SKLinear computes the average over \texttt{num\_terms} sketched matrix multiplication operations where the rank of each sketch matrix is specified by \texttt{low\_rank}. By increasing \texttt{num\_terms}, we get results that are closer to the expected value at the cost of increasing the number of parameters. This follows the proposed sketching mathematical formulation from \cite{kasiviswanathan2017deepneuralnetworkapproximation}. Listing \ref{lst:panther-example} demonstrates this replacement pattern:

\begin{lstlisting}[style=acmcode, language=Python, caption={Using Panther as a drop-in replacement for a PyTorch layer.}, label={lst:panther-example}]
import torch.nn as nn
import panther as pr

# Standard PyTorch model
class StandardModel(nn.Module):
    def __init__(self):
        super().__init__()
        self.fc1 = nn.Linear(8192, 8192)

# Panther-optimized model (2-3x speedup on example hyperparameters)
class PantherModel(nn.Module):
    def __init__(self):
        super().__init__()
        self.fc1 = pr.nn.SKLinear(8192, 8192, num_terms=1, low_rank=16)
\end{lstlisting}

\subsection{After Development use-case}

Panther makes it easy to migrate to randomized layers even after development using the SKAutoTuner, which automates the tedious process of navigating model hierarchies, selecting layers, and discovering optimal sketching parameters. Listing \ref{lst:tuner-example} demonstrates automatic optimization of a pre-trained BERT model by targeting all Linear layers in the transformer encoder and optimizing for speed while maintaing a quality metric constraint (e.g., Masked Language Modeling (MLM) loss):

\begin{lstlisting}[style=acmcode, language=Python, caption={Easy  Migration with SKAutoTuner.}, label={lst:tuner-example}]
from transformers import BertForMaskedLM
from panther.tuner import SKAutoTuner, LayerConfig, TuningConfigs

# Load pre-trained BERT
model = BertForMaskedLM.from_pretrained("bert-base-uncased")

# Configure automatic layer discovery and tuning
config = LayerConfig(
    layer_names={"type": "Linear"},
    params="auto",  # Automatic search space
    separate=True,  # Optimize each layer independently
    copy_weights=True  # Preserve trained weights
)

# Create tuner with a quality metric constraint
tuner = SKAutoTuner(
    model=model,
    configs=TuningConfigs([config]),
    accuracy_eval_func=eval_quality,
    accuracy_threshold=thresh, # Based on eval_quality
    optmization_eval_func=speed_eval_func,
    search_algorithm=OptunaSearch(n_trials=10)
)

# Search and apply optimal configuration
tuner.tune()
optimized_model = tuner.apply_best_params()
\end{lstlisting}

\section{Evaluation}
\label{sec:eval}

\subsection{Runtime and Memory}

To characterize the runtime and memory behavior of Panther's sketched operators, we performed a comprehensive set of benchmarks\footnote{\textbf{All benchmarks}, documentation, and examples are available at \cite{docs}.} spanning fully connected layers, convolutional layers, and attention mechanisms. For each module, we measured forward and backward pass runtime, reported as the mean over 200 repeated trials, as well as peak memory consumption during execution for multi-head attention. For linear and convolution layers, the reduction in layer size can be computed analytically \cite{kasiviswanathan2017deepneuralnetworkapproximation}. Experiments were conducted on NVIDIA Tesla T4 and P100 GPUs, enabling evaluation across hardware generations. Results were compared against established baselines, including PyTorch's \texttt{nn.Linear}, \texttt{nn.Conv2d}, and \texttt{nn.MultiheadAttention}.

For the sketched fully connected layers (SKLinear), we varied input and output dimensions $d_{\text{in}}, d_{\text{out}} \in \{256,\allowbreak 512, 1024,\allowbreak 8192,\allowbreak 16384,\allowbreak 32768,\allowbreak 65536\}$, the number of sketch terms $l \in \{1,\allowbreak 2,\allowbreak 3\}$, and the target low-rank dimension $k \in \{16,\allowbreak 32,\allowbreak 64,\allowbreak 128,\allowbreak 256,\allowbreak 512\}$. These parameters directly control the approximation rank and expressive capacity of the layer, trading accuracy for computational and memory efficiency. To ensure fair comparisons, benchmarks were skipped whenever the sketched parameterization exceeded the original layer size, i.e., $2 l k (d_{\text{in}} + d_{\text{out}}) > d_{\text{in}} d_{\text{out}}$, as shown in \cite{kasiviswanathan2017deepneuralnetworkapproximation}, since such configurations cannot yield theoretical speedups.

\begin{figure}[h]
  \centering
  \includegraphics[width=\linewidth]{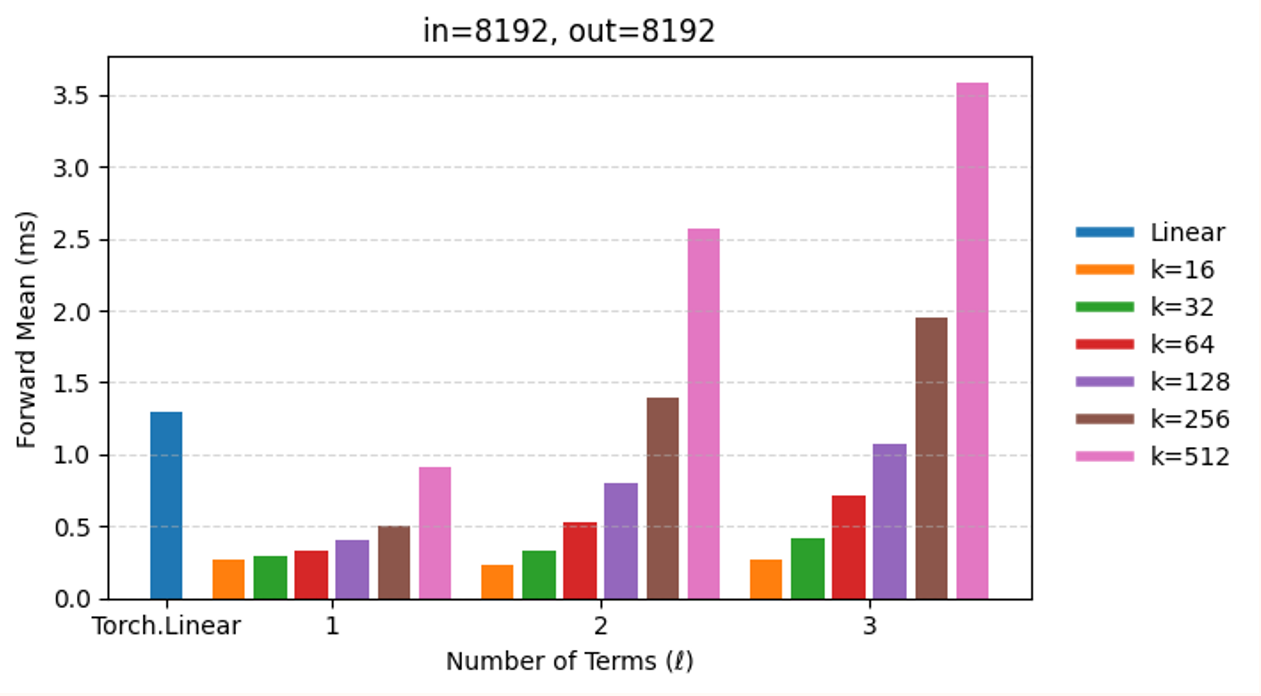}
  \caption{Forward pass runtime (ms) for the sketched Linear layer \cite{kasiviswanathan2017deepneuralnetworkapproximation} compared to PyTorch. Run is for input and output features of 8192 and varies the introduced hyperparameters number of terms ($l$) and low rank ($k$)}
  \Description{Benchmark for forward pass varying the hyperparameters and demonstrating a speedup depending on parameters}
  \label{fig:forward-linear}
\end{figure}

Similarly, for the sketched convolutional layers (SKConv2D), we evaluated square kernels of sizes $3, 5, 9$, input image resolutions $\{64,\allowbreak 128,\allowbreak 256\}$, channel dimensions ranging from 64 to 2048, and sketch parameters $l \in \{1,\allowbreak 2,\allowbreak 3\}, k \in \{8,\allowbreak 16,\allowbreak 32\}$. Larger kernels and channel counts amplify the cost of dense convolution, making them particularly suitable for low-rank sketching and allowing us to study how approximation structure impacts memory bandwidth and compute intensity.

\begin{figure}[h]
  \centering
  \includegraphics[width=\linewidth]{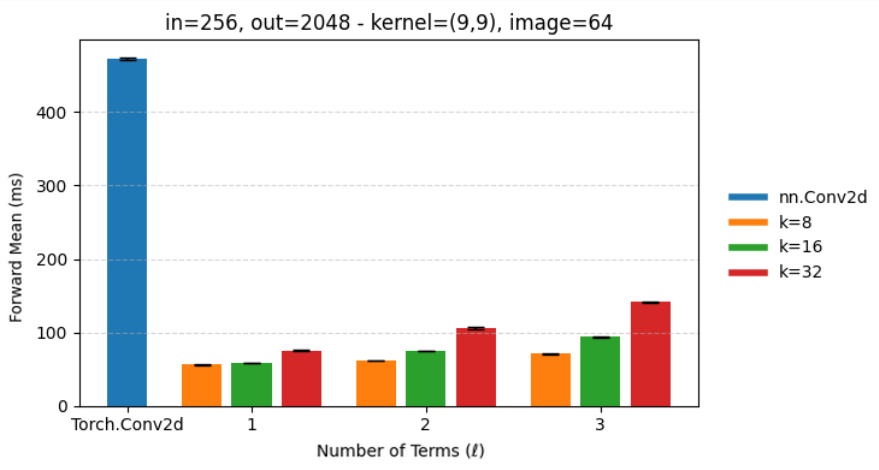}
  \caption{Forward pass runtime (ms) for the sketched Conv2D layer \cite{kasiviswanathan2017deepneuralnetworkapproximation} compared to PyTorch. Run is for input and output channels of $256 \times 2048$ with a squared kernel and image of size $9$ and $64$ respectively. We vary the introduced hyperparameters number of terms ($l$) and low rank ($k$)}
  \Description{Benchmark for forward pass varying the hyperparameters and demonstrating a speedup depending on parameters}
  \label{fig:forward-conv}
\end{figure}

Finally, for Performers, we benchmarked embedding dimensions $\{128, 256, 512, 1024\}$, with head counts $\{4,\allowbreak 8,\allowbreak 16\}$, random feature dimensions $\{64,\allowbreak 128,\allowbreak 256\}$, kernel functions $\{\text{Softmax},\allowbreak \text{ReLU}\}$, and sequence lengths of up to 8192 tokens. These parameters determine the fidelity of the random-feature approximation and directly influence both quadratic attention costs and memory footprint.

\begin{figure}[h]
  \centering
  \includegraphics[width=\linewidth]{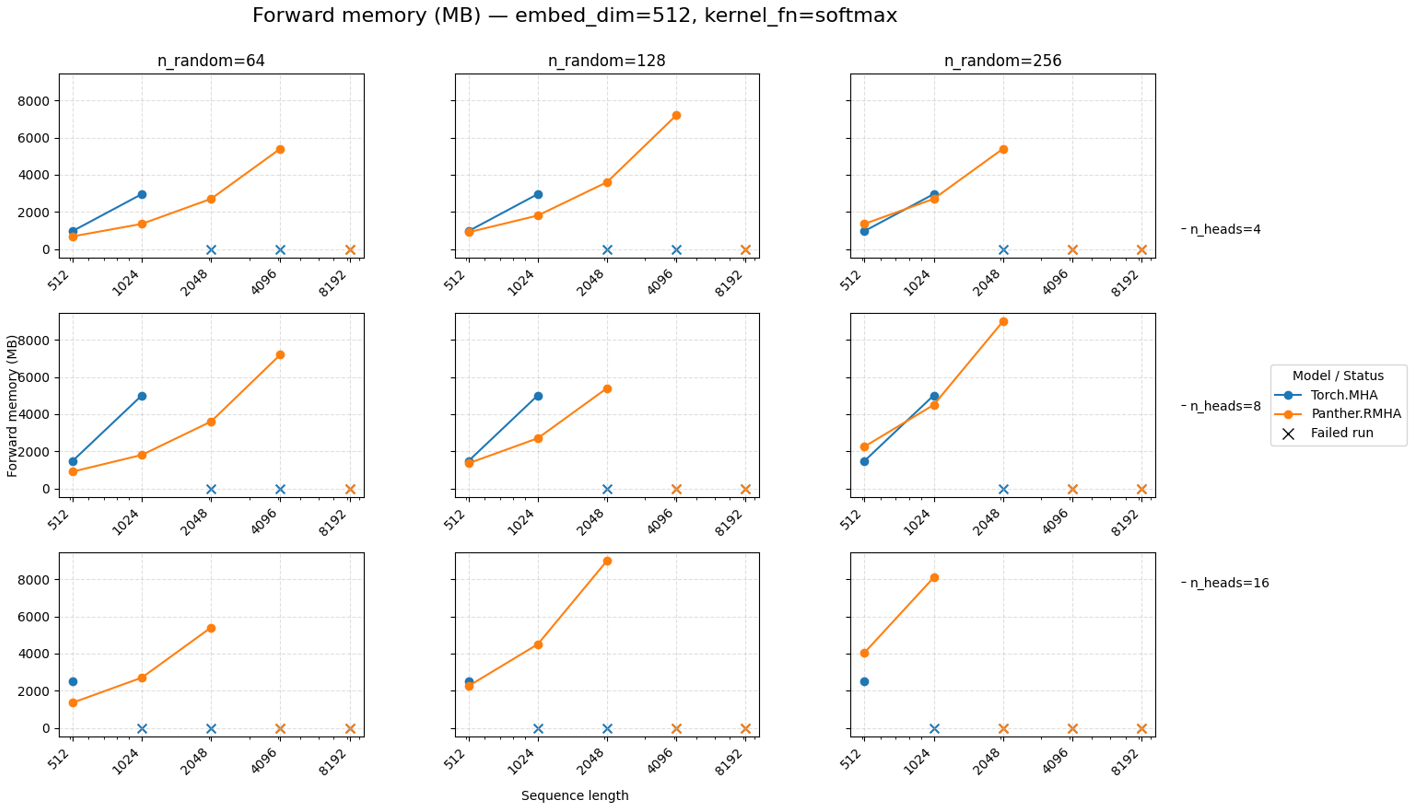}
  \caption{Forward pass memory (MB) comparison for Panther's RandMultiHeadAttention using linear attention of Performers \cite{choromanski2022rethinkingattentionperformers} compared to PyTorch. Run is for embed dimension of 512 using a softmax kernel and varies sequence length, number of heads, and the introduced random features hyperparameter.}
  \Description{Forward pass memory footprint showing superiority due to linearized attention mechanism.}
  \label{fig:forward-attention}
\end{figure}

Figures~\ref{fig:forward-linear},\ref{fig:forward-conv} and~\ref{fig:forward-attention} illustrate the practical benefits and trade-offs of Panther's layers. Figure~\ref{fig:forward-linear} shows the forward pass runtime (in milliseconds) for the sketched fully connected layer (SKLinear) \cite{kasiviswanathan2017deepneuralnetworkapproximation} with $d_{\text{in}} = d_{\text{out}} = 8192$, varying the number of terms $\ell$ and the low-rank dimension $k$. Compared to PyTorch's dense \texttt{nn.Linear} layer, smaller values of $k$ achieve substantial speedups, particularly for $\ell = 1$ or $2$, while larger $k$ approaches and exceeds the cost of the dense baseline.

Figure~\ref{fig:forward-conv} illustrates the runtime (in milliseconds) of the sketched 2D convolution layer (SKConv2d) compared to the standard implementation. The results demonstrate the efficiency of the sketching method that across all tested settings, SKConv2d significantly outperforms PyTorch's \texttt{nn.Conv2d}, achieving substantially lower forward pass latencies.

Figure~\ref{fig:forward-attention} reports peak forward memory usage for Performer \cite{choromanski2022rethinkingattentionperformers} with embedding dimension $512$ and a softmax kernel, varying sequence length, number of heads, and number of random features. Notably, Panther successfully executes configurations where PyTorch fails due to memory limits (indicated by ``$\times$'' markers), demonstrating the extended range of feasible sequence lengths.

Together, these results highlight Panther's ability to implement the significant speed and memory advantages we see in literature that enable larger workloads that are infeasible with standard implementations.

\subsection{Quality}

To verify that memory savings do not come at the cost of model utility, we evaluated Panther with the SKAutoTuner to find the best parameters using the WikiText\cite{merity2016pointer} dataset and MLM loss on BERT \cite{bertModel} model. We replaced the dense linear layers within the model with Panther's \texttt{SKLinear} equivalents. The results demonstrate that the model achieves up to $75\%$ reduction in size while maintaining a comparable MLM loss value ($4.601$ and $4.594$). Crucially, Panther facilitates this transition with minimal engineering overhead; as a library designed to implement existing sketching literature, it allows users to perform these optimizations with just a few lines of code as seen in Section~\ref{sec:tool}, removing the requirement for deep expertise in RandNLA or extra learning overhead beyond normal PyTorch workflows.

We further demonstrated the library's versatility through a case study on the ResNet-50 \cite{he2016deep} model. By utilizing Panther to replace standard 2D convolution layers at a controlled size reduction of $30\%$, we observed a marginal accuracy decrease from $89\%$ to $86\%$ on the CIFAR-10 \cite{Krizhevsky09learningmultiple} dataset. This confirms that Panther can be easily adapted to different model architectures, streamlining the application of RandNLA as a compression technique.

\section{Related Work}
\label{sec:related}

The intersection of RandNLA and deep learning has evolved from theoretical approximations to practical software ecosystems. At the primitive level, libraries like \textbf{RandBLAS} and \textbf{RandLAPACK} have established standard C++ interfaces for sketching operations~\cite{murray2023randomizednumericallinearalgebra}, recently expanding to GPU-accelerated implementations~\cite{shah2025kokkos}. Specific advances in matrix decomposition, such as the \textbf{CQRRPT} algorithm (CholeskyQR with Randomization and Pivoting), have proven critical for stable computations on tall matrices~\cite{melnichenko2025choleskyqrrandomizationpivotingtall}. In the domain of structured matrices, \textbf{Compositional Linear Algebra (CoLA)}~\cite{potapczynski2023cola} automates efficient operations for matrices with compositional structure (e.g., Kronecker products), though primarily within the JAX ecosystem. For neural networks, \textbf{Tensor Sketching} has been applied to approximate convolutional layers~\cite{kasiviswanathan2017deepneuralnetworkapproximation}, with recent extensions like \textbf{CTSketch} enabling scalable neurosymbolic learning~\cite{shah2025ctsketch}. Similarly, randomized Singular Value Decomposition (\textbf{RSVD}) remains a cornerstone for model compression, with contemporary approaches integrating layer-wise rank selection directly into the training loop~\cite{yang2025integrating}. Panther distinguishes itself by unifying these disparate C++ primitives and sketching algorithms into a cohesive PyTorch-native library, abstracting the low-level complexity of these methods behind standard \texttt{nn.Module} interfaces.

\section{Conclusion}
\label{sec:conclusion}
Panther represents the first production-grade library to bring the theoretical benefits of RandNLA to the PyTorch community. By integrating robust, well-tested sketching primitives and a native backend, Panther serves as a direct integrator for RandNLA techniques to relieve the compute bottleneck in large-scale linear algebra and neural-network workloads, enabling orders-of-magnitude reductions in working memory and time while allowing practitioners to train and evaluate larger models or larger batches on the same hardware with minimal and controllable sacrifice in accuracy.

Future development will focus on expanding the catalog of sketching operators, extension of RandNLA techniques to other deep learning layers in PyTorch, as well as providing and testing builds for multiple software and packaging targets (PyPI wheels, conda packages, and platform-specific binaries). We welcome community contributions, issue reports, and pull requests to help grow the library and its ecosystem.

\begin{acks}

We would like to extend thanks to Eyad Salama and Muhammad ElNokrashy for their valuable feedback and discussions throughout the development of this work.
\end{acks}

\bibliographystyle{ACM-Reference-Format}
\bibliography{refs}



\end{document}